\documentclass{article}

    \PassOptionsToPackage{numbers, compress}{natbib}
\usepackage[preprint]{neurips_2026}

\usepackage[utf8]{inputenc} 
\usepackage[T1]{fontenc}    
\usepackage{hyperref}       
\usepackage{url}            
\usepackage{booktabs}       
\usepackage{amsfonts}       
\usepackage{nicefrac}       
\usepackage{microtype}      
\usepackage{xcolor}         
\usepackage{tcolorbox}
\usepackage{graphicx}
\usepackage{amsmath}
\usepackage{amsthm}
\usepackage{multirow}
\usepackage{wrapfig}
\usepackage{enumitem}

\title{Adding Thermal Awareness to Visual Systems in Real-Time via Distilled Diffusion Models}

\author{
  Yuchen Guo$^{1,}$\thanks{Correspondence to: yuchenguo2027@u.northwestern.edu, wfsu@bnbu.edu.cn.} \quad
  Junli Gong$^{2}$ \quad
  Wenjun Dong$^{1}$ \quad
  Yiuming Cheung$^{3}$ \quad
  Weifeng Su$^{4,}$\footnotemark[1] \\ \\
  $^1$Northwestern University \quad
  $^2$Northeastern University \\
  $^3$Hong Kong Baptist University \quad
  $^4$Beijing Normal - Hong Kong Baptist University
}

\begin{document}

\maketitle



\vspace{-14pt}

\begin{abstract}
Purely RGB-based vision models often fail to provide reliable cues in challenging scenarios such as nighttime and fog, leading to degraded performance and safety risks. Infrared imaging captures heat-emitting sources and provides critical complementary information, but existing high-fidelity fusion methods suffer from prohibitive latency, rendering them impractical for real-time edge deployment. To address this, we propose \textbf{\textit{FusionProxy}}, a real-time image fusion module designed as a fully independent, plug-and-play component with diffusion level quality. FusionProxy exploits two complementary statistics of a teacher sample ensemble: per-pixel variance in raw image space, used to weight pixel-level supervision, and per-pixel variance inside frozen foundation backbones, used to route feature-level alignment spatially. Once trained, FusionProxy can be directly integrated into any visual perception system without joint optimization. Extensive experiments demonstrate that our method achieves superior performance on static recognition tasks and significantly enhances robustness in dynamic tasks, including closed-loop autonomous driving. Crucially, FusionProxy achieves real-time inference speeds on diverse platforms, from high-end GPUs to commodity hardware, providing a flexible and generalizable solution for all-day perception. The source code will be available.
\end{abstract}
\vspace{-23pt}

\begin{center}
\includegraphics[
    width=\textwidth,
    height=0.43\textheight,
    keepaspectratio
]{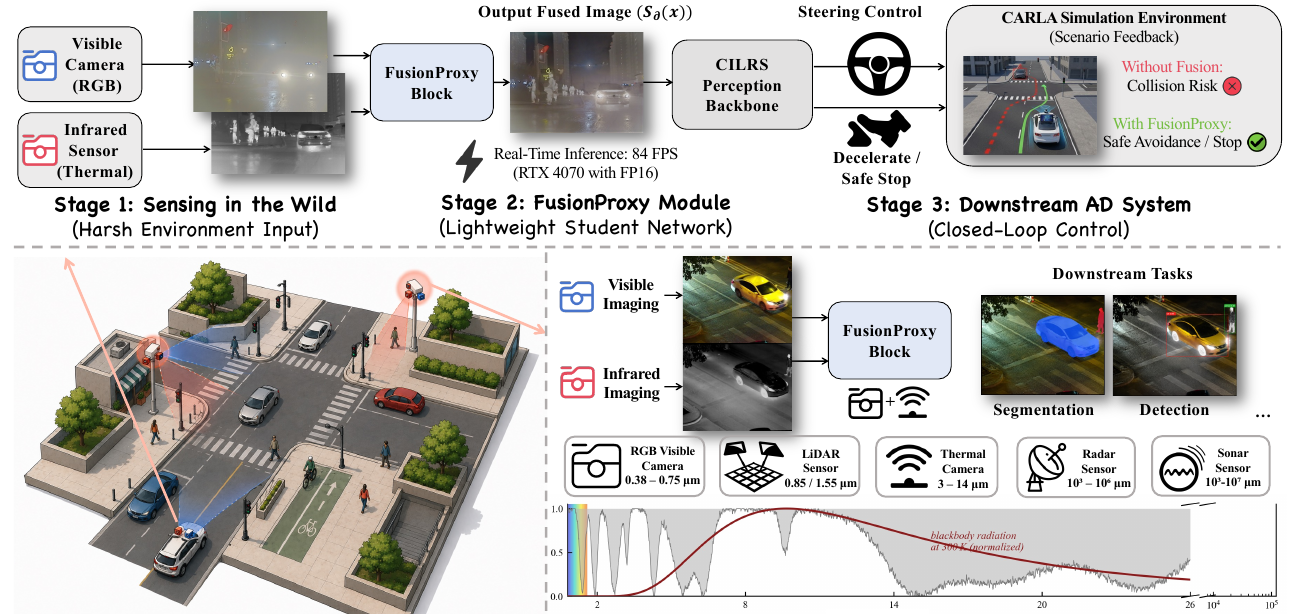}
\end{center}

\vspace{-10pt}

{\small
\noindent Figure 1:
\textit{Top:} End-to-end closed-loop autonomous driving pipeline under degraded visibility.
\textit{Bottom-left:} Thermal radiation exposes pedestrians and vehicles invisible to RGB.
\textit{Bottom-right:} The fused output is directly consumable by frozen downstream models without retraining.
}

\vspace{-10pt}

\setcounter{figure}{1}

\section{Introduction}
\label{sec:intro}

Human visual perception, and most of the machine vision we have built (\textit{e.g.}, autonomous vehicles and surveillance), operates within the visible spectrum (\textit{i.e.}, RGB imaging), which leaves both vulnerable in conditions of degraded illumination such as nighttime, fog and glare. To compensate, modern perception systems augment RGB cameras with active sensors such as LiDAR, radar, and sonar, which emit signals and measure their return. Active sensing provides  additional information about the surrounding scene for situational awareness, but it has a fundamental scalability limitation: emitted signals interfere with one another as the density of intelligent embodied agents scales up~\citep{verghese2017self, popko2020interference}; cost, power, and form-factor budgets per unit further restrict deployment. A scalable substitute should therefore be \emph{passive}, drawing information from signals that already exist in the environment rather than ones it generates \citep{bao2023heat}.

Thermal infrared radiation is the natural candidate. Every 
object above absolute zero emits thermal radiation, making 
heat an omnipresent signal that requires no transmission, no 
synchronization, and no coordination across agents. Thermal infrared is also uniquely suited to integration with 
the existing visual stack. Unlike LiDAR's point clouds, 
radar's range-Doppler maps, or event cameras' asynchronous 
spike streams, all of which demand new processing pipelines, 
thermal cameras produce a 2D pixel grid, the same format as 
an RGB image. Image fusion \citep{archana2024deep, Zhao_2024_emma} is the natural mechanism that 
exploits this shared format alignment, it aims to produce a single fused image that combines the thermal information of infrared image $I_{\mathrm{IR}}$ with the structural and textural information of visible image $I_{\mathrm{VIS}}$~\citep{ma2019infrared, 
zhang2021image}. Because the output remains an RGB-format image, it serves a dual role: \emph{machine-side}, it is consumed as a drop-in replacement for $I_{\mathrm{VIS}}$ by any frozen RGB-pretrained model, including detectors, segmenters, vision-language models, and driving policies; \emph{human-side}, it is directly viewable on the same displays, dashboards, and monitors that already mediate human-RGB interaction. This dual-use property is what makes image fusion the right integration mechanism for adding thermal awareness to the existing RGB-based visual stack.

Despite this conceptual fit, no current image fusion method 
makes the integration practical. The frontier of fusion 
quality is occupied by diffusion-based methods such as 
DDFM~\citep{zhao2023ddfm}, 
Mask-DiFuser~\citep{tang2025mask}, 
Text-IF~\citep{yi2024text}, and ControlFusion~\citep{tang2025controlfusion}, which produce 
visually faithful fusions through iterative sampling. 
Iterative sampling, however, places these methods at one 
frame every several seconds on server-class GPUs. Although one-step diffusion~\citep{wang2025efficient} minimizes inference steps, the distilled generator retains the heavy architecture of original diffusion models, which is incompatible with any real-time perception pipeline. At the opposite end, real-time methods such as TarDAL~\citep{liu2022target} sacrifice fusion quality 
substantially, falling well behind the diffusion-quality fusions that motivate the integration in the first place. We therefore ask: \textbf{\textit{can we combine the fusion quality of diffusion models with the inference speed required for commodity deployment?}}

We answer this question with \textbf{FusionProxy}, a plug-and-play and real-time fusion module with respect to downstream perception models. On the one hand, we construct a diffusion teacher ensemble by drawing multiple samples from two complementary pre-trained diffusion fusion models per training image, yielding per-pixel sample statistics in both image space and foundation feature space. On the other hand, we introduce a dual-signal distillation loss that turns these statistics into supervision: image-space variance weights pixel-level matching against the ensemble mean, while feature-space variance routes feature-level alignment across a panel of frozen foundation backbones, so that each region is supervised by the backbone locally most informative about it. Based on the above, we distil the ensemble into a deterministic single-pass student whose architecture is decoupled from the teacher's denoising network, allowing lightweight backbones to inherit diffusion-grade fusion quality. Once trained, FusionProxy can be deployed alongside an infrared sensor as a perception front-end, enabling immediate integration with downstream visual systems without requiring any retraining. Our contributions are as follows:

\begin{itemize}
\setlength\itemsep{0.2em}
\item We propose \textbf{FusionProxy}, a plug-and-play fusion module that distills a diffusion teacher into a lightweight student, achieving diffusion-level fusion quality under strict real-time constraints on commodity hardware.

\item We introduce a dual-signal distillation framework in which a single diffusion teacher ensemble simultaneously drives uncertainty-weighted pixel supervision (via image-space sample variance) and spatially-adaptive multi-foundation alignment, with both signals derived from a single set of cached teacher forward passes.

\item We validate FusionProxy through extensive image fusion benchmarks, plug-and-play deployment on frozen RGB-pretrained detection and segmentation models, and an end-to-end closed-loop autonomous driving pipeline.
\end{itemize}

\section{Related Works}
\label{app:related}
\textbf{Image Fusion.} Image Fusion aims to integrate complementary information from different imaging modalities into a single image \citep{Zhao_2024_emma, zhang2021image, archana2024deep, guo2024fuse4seg}. Autoencoder based methods focus on the elaborated reconstruction and fusion loss functions \cite{zhao2020didfuse,Zhao_2023_cddfuse, guo2024dae}, while GAN-based image fusion methods aim to contrast two discriminators for fused images and both source images\citep{ma2020ddcgan,liu2022target}. Recently, a few studies have explored incorporating diffusion models into image\cite{zhao2023ddfm,yi2024diff,tang2025controlfusion} and introduced video fusion\cite{guo2024dae,zhao2025unified}. However, existing approaches predominantly focus on static image fusion or handle video fusion with prohibitively slow inference, limiting their practical usability. In contrast, we aim for real-time video fusion with plug-and-play deployment capability.

\textbf{Diffusion Models.} Recently, diffusion models have achieved amazing performance across a broad range of tasks, including image synthesis\cite{zhang2023adding}, video generation\cite{zhang2024mimicmotion,peebles2023scalable}, and image restoration\cite{lin2025harnessing,yu2024scaling}. Moreover, recent efforts on accelerating diffusion models through distillation\cite{meng2023distillation,song2023consistency} have significantly improved their practicality. One-step diffusion for image fusion model\cite{wang2025efficient} reduces the number of inference steps, but the distilled generator still inherits the computationally heavy architecture, resulting in non-negligible latency that hinders real-time deployment.

\section{Methodology}
\label{sec:method}

\subsection{Problem Formulation}
\label{sec:problem_setup}

Let $x = (I_{\mathrm{IR}}, I_{\mathrm{VIS}}) \in \mathcal{X}$ denote an aligned infrared-visible input pair, where $I_{\mathrm{IR}} \in \mathbb{R}^{H \times W}$ and $I_{\mathrm{VIS}} \in \mathbb{R}^{H \times W \times 3}$. A pre-trained diffusion fusion model $T$ defines a conditional implicit distribution $p_T(y \mid x)$ over fused images $y \in \mathcal{Y} \subset \mathbb{R}^{H \times W \times 3}$, from which high-quality samples $y_T \sim p_T(y \mid x)$ can be drawn via iterative denoising, at a cost prohibitive for real-time deployment.

Our goal is to learn a lightweight, deterministic student $S_\theta : \mathcal{X} \to \mathcal{Y}$ that approximates the teacher in a single forward pass:
\begin{equation}
\hat{y} \;=\; S_\theta(x) \;\approx\; \mathbb{E}_{y \sim p_T(y \mid x)}[y].
\label{eq:student_target}
\end{equation}
The expectation form encodes the deterministic-student requirement: a single input must map to a single, stable output, so that $\hat{y}$ is consumable by frozen RGB-pretrained downstream models $\mathcal{F}$ in place of $I_{\mathrm{VIS}}$. We further require $S_\theta$ to run at $r_{\min} \geq 30$ FPS on commodity hardware.

Direct optimization of Eq.~(\ref{eq:student_target}) is infeasible: the expectation has no closed form. We instead approximate it via Monte Carlo by drawing $\{y_T^{(n)}\}_{n=1}^{N} \overset{\text{iid}}{\sim} p_T(y \mid x)$ and minimizing
\begin{equation}
\theta^{\star} \;=\; \arg\min_\theta \; \mathbb{E}_x \; \mathcal{L}\bigl( S_\theta(x),\, \{y_T^{(n)}\}_{n=1}^{N} \bigr).
\label{eq:distillation_obj}
\end{equation}
Section~\ref{sec:teacher_ensemble} instantiates the teacher and Section~\ref{sec:method_loss} develops $\mathcal{L}$.

\subsection{Diffusion Teacher Ensemble}
\label{sec:teacher_ensemble}

We instantiate the teacher in Eq.~(\ref{eq:distillation_obj}) as an ensemble drawn from two pre-trained diffusion fusion models, addressing the bias of any single teacher and providing the structured statistics that drive our distillation loss in Sec.~\ref{sec:method_loss}.

\noindent\textbf{Dual teachers.}
The single-teacher formulation in Eq.~(\ref{eq:distillation_obj}) is straightforwardly extended to multiple teachers by drawing samples from each in turn. We use two diffusion teachers with complementary training distributions. \emph{DDFM}~\citep{zhao2023ddfm} is a Bayesian conditional diffusion model trained directly on IR-VIS pairs and provides domain-grounded thermal radiation modeling. \emph{Mask-DiFuser}~\citep{tang2025mask} is a modality-agnostic model trained on natural images via dual masked restoration and provides broad image priors and texture fidelity. Distilling from a single teacher would inherit that teacher's distributional bias; the two together cover complementary failure modes.

\noindent\textbf{Sampling protocol.}
For each input $x = (I_{\mathrm{IR}}, I_{\mathrm{VIS}})$ we draw $N$ DDIM samples~\citep{song2020denoising} from each teacher, yielding the ensemble
\begin{equation}
\mathcal{Y}_T(x) \;=\; \bigl\{ y_T^{(n)} \bigr\}_{n=1}^{2N}, \qquad y_T^{(n)} \sim p_{T(n)}(y \mid x),
\end{equation}
where $T(n) \in \{\text{DDFM}, \text{Mask-DiFuser}\}$ indexes the source teacher of the $n$-th sample. We use $N = 4$ samples per teacher throughout (ablation in Sec.~\ref{sec:exp_ablation}).

\begin{figure*}[t!]
    \centering
    \includegraphics[width=\textwidth]{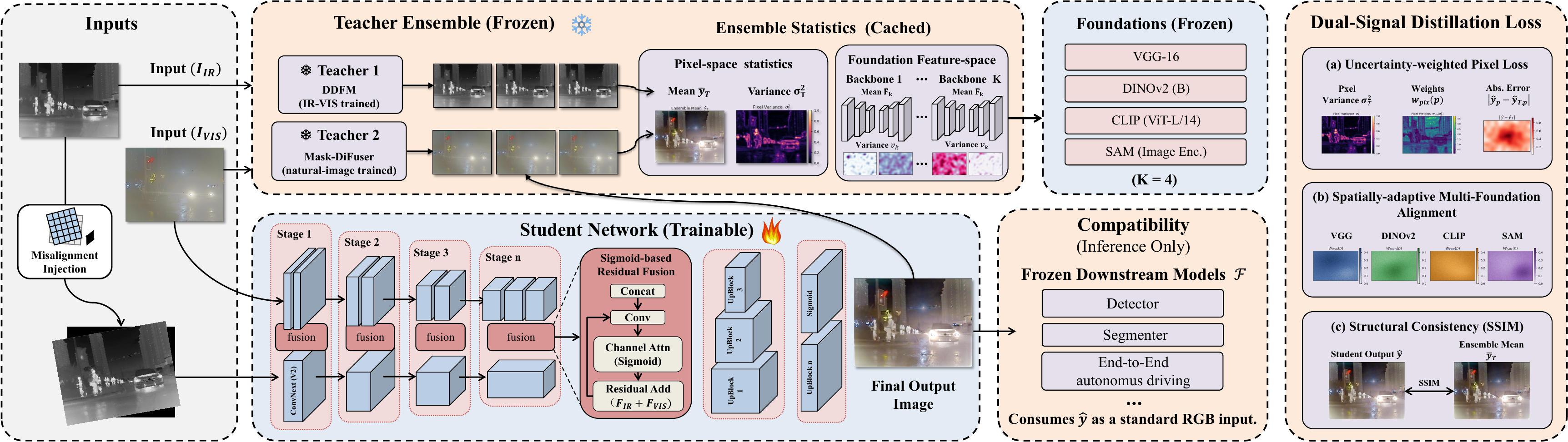}
    \caption{The FusionProxy framework. Dual diffusion teachers 
    generate a sample ensemble whose mean and per-pixel 
    variance drive uncertainty-weighted pixel supervision. 
    The student is aligned to a panel of frozen foundation 
    models via spatially-adaptive weights derived from per-backbone informativeness.}
    \label{fig:method}
    \vspace{-9pt}
\end{figure*}

\noindent\textbf{Ensemble statistics.}
The pointwise ensemble mean
\begin{equation}
\bar{y}_T(x) \;=\; \frac{1}{2N} \sum_{n=1}^{2N} y_T^{(n)}(x)
\label{eq:ensemble_mean}
\end{equation}
serves as a low-variance Monte Carlo estimate of the teacher expectation $\mathbb{E}_{y \sim p_T(y \mid x)}[y]$ targeted in Eq.~(\ref{eq:student_target}). The per-pixel sample variance
\begin{equation}
\sigma_T^2(x, p) \;=\; \mathrm{Var}_n\bigl[ y_T^{(n)}(x, p) \bigr]
\label{eq:pixel_var}
\end{equation}
quantifies the ambiguity of this estimate at pixel $p$: high variance flags pixels where the teacher distribution is multimodal in raw image space.

In addition to image-space statistics, we measure the same ensemble inside frozen foundation backbones $\{\Phi_k\}_{k=1}^{K}$. Let $\Phi_k(y_T^{(n)})_p$ denote the feature of $\Phi_k$ at spatial location $p$ for the $n$-th teacher sample. The per-backbone, per-pixel feature variance is
\begin{equation}
v_k(x, p) \;=\; \mathrm{Var}_n\bigl[ \Phi_k(y_T^{(n)})_p \bigr].
\label{eq:feat_var}
\end{equation}
This quantity captures how strongly backbone $\Phi_k$ responds to fusion ambiguity at pixel $p$, and serves as the routing signal for multi-foundation alignment in Sec.~\ref{sec:method_loss}.

\noindent\textbf{Caching.}
All teacher samples $\mathcal{Y}_T(x)$ and foundation features $\{\Phi_k(y_T^{(n)})\}$ are precomputed once per training image and cached, since both teachers and foundation backbones are frozen. Teacher inference and foundation extraction are therefore one-time amortized costs and do not appear in the per-iteration training budget.

\subsection{Dual-Signal Distillation Loss}
\label{sec:method_loss}

The teacher ensemble defined in Sec.~\ref{sec:teacher_ensemble} provides a single pseudo-target $\bar{y}_T$ together with two complementary statistics: the pixel-space variance $\sigma_T^2$ and the feature-space variance $v_k$. We use the first to weight pixel-level supervision, the second to route feature-level supervision across multiple foundation backbones. Both signals are derived from the same cached forward passes through the teachers and incur no additional inference at training time.

\noindent\textbf{Uncertainty-weighted pixel supervision.}
The pseudo-target $\bar{y}_T$ is reliable where the teacher ensemble agrees and unreliable where it disagrees. Forcing the student to match $\bar{y}_T$ uniformly would commit it to teacher noise in high-variance regions. We therefore weight pixel-level supervision by the inverse of the pixel-space variance $\sigma_T^2(x, p)$ from Eq.~(\ref{eq:pixel_var}):
\begin{equation}
\mathcal{L}_{\mathrm{pix}}
\;=\;
\sum_{p} w_{\mathrm{pix}}(x, p) \cdot \bigl\| S_\theta(x)_p - \bar{y}_T(x)_p \bigr\|_1,
\qquad
w_{\mathrm{pix}}(x, p) \;=\; \frac{ 1 / \bigl( \sigma_T^2(x, p) + \epsilon \bigr) }{ \sum_{p'} 1 / \bigl( \sigma_T^2(x, p') + \epsilon \bigr) },
\label{eq:pix_loss}
\end{equation}
where $\epsilon = 10^{-3}$ stabilizes the inverse and the weights are normalized over the spatial domain $\Omega$ so that $\sum_p w_{\mathrm{pix}}(x, p) = 1$. Pixels where the teacher ensemble is internally consistent contribute proportionally more gradient; pixels where it is multimodal contribute less.
  
\noindent\textbf{Spatially-adaptive multi-foundation alignment.}
Compatibility with downstream RGB-pretrained models (Sec.~\ref{sec:problem_setup}) requires that $S_\theta(x)$ remain consistent with the teacher ensemble in the feature spaces those models use. Since these feature spaces are heterogeneous, we align the student to a panel of $K$ frozen foundation backbones $\{\Phi_k\}_{k=1}^{K}$: VGG-16~\citep{simonyan2014very}, DINOv2~\citep{oquab2023dinov2}, CLIP~\citep{radford2021learning}, and SAM image encoder~\citep{kirillov2023segment}. All foundation features are bilinearly resampled to a common spatial grid $\Omega$ and rescaled per-channel to unit variance: $\tilde{\Phi}_k = \Phi_k / \hat{\sigma}_k$, with $\hat{\sigma}_k$ the channel-wise standard deviation of $\Phi_k$ over the training set.

A uniform sum across $\{\Phi_k\}$ is dominated by whichever backbone has the largest local response, and ignores that different backbones are informative in different regions: SAM near object boundaries, CLIP in semantically rich regions, VGG in textured areas, DINOv2 in mid-level structural regions. We therefore weight each backbone's contribution at each pixel by its local informativeness, measured as its feature variance across teacher samples. To prevent backbones with globally larger feature variance from dominating, we normalize each backbone's variance by its training-set mean before computing routing weights:
\begin{equation}
\tilde{v}_k(x, p) \;=\; \frac{ v_k(x, p) }{ \mathbb{E}_{x'} \bigl[ \overline{v_k}(x') \bigr] + \epsilon },
\qquad
W_k(x, p) \;=\; \frac{ \exp\bigl( \tilde{v}_k(x, p) / \tau \bigr) }{ \sum_{k'} \exp\bigl( \tilde{v}_{k'}(x, p) / \tau \bigr) },
\label{eq:routing}
\end{equation}
where $\overline{v_k}(x') = \frac{1}{|\Omega|} \sum_p v_k(x', p)$ is the spatial mean of $v_k$ on image $x'$, and $\tau$ is a temperature controlling routing sharpness. Setting $\tau \to 0$ assigns each pixel to its single most informative backbone; $\tau \to \infty$ recovers uniform weighting. We use $\tau = 1.0$ throughout.
The alignment target in foundation feature space is computed as the mean of features extracted from individual teacher samples, which preserves high-frequency feature responses that are smoothed out by averaging in pixel space:
\begin{equation}
\bar{F}_k(x)_p \;=\; \frac{1}{2N} \sum_{n=1}^{2N} \tilde{\Phi}_k\bigl( y_T^{(n)}(x) \bigr)_p.
\label{eq:feat_target}
\end{equation}
The multi-foundation alignment loss is then
\begin{equation}
\mathcal{L}_{\mathrm{MFM}}
\;=\;
\frac{1}{|\Omega|} \sum_{k=1}^{K} \sum_{p \in \Omega}
W_k(x, p) \cdot
\bigl\| \tilde{\Phi}_k(S_\theta(x))_p - \bar{F}_k(x)_p \bigr\|_2^2.
\label{eq:mfm_loss}
\end{equation}
\noindent\textbf{Structural consistency.}
Pixel and feature losses against the ensemble mean $\bar{y}_T$ are vulnerable to over-smoothing, since $\bar{y}_T$ averages out high-frequency details that vary across teacher samples. We add a structural term based on SSIM~\citep{wang2004ssim}, which is computed over local windows and is invariant to such averaging in its low-order statistics:
\begin{equation}
\mathcal{L}_{\mathrm{ssim}}
\;=\;
1 - \mathrm{SSIM}\bigl( S_\theta(x),\, \bar{y}_T(x) \bigr).
\label{eq:ssim_loss}
\end{equation}

\noindent\textbf{Total objective.}
The full training loss is a weighted sum of the three terms:
\begin{equation}
\mathcal{L} \;=\;
\lambda_{\mathrm{pix}} \mathcal{L}_{\mathrm{pix}}
\;+\; \lambda_{\mathrm{MFM}} \mathcal{L}_{\mathrm{MFM}}
\;+\; \lambda_{\mathrm{ssim}} \mathcal{L}_{\mathrm{ssim}}.
\label{eq:total_loss}
\end{equation}
We use $\lambda_{\mathrm{pix}} = 1.0$, $\lambda_{\mathrm{MFM}} = 0.5$, $\lambda_{\mathrm{ssim}} = 0.2$ throughout. Only $S_\theta$ is updated during training; teachers, foundation backbones, and all cached statistics are frozen. At inference time, only $S_\theta$ runs; teachers, foundation backbones, and the cached statistics are not invoked.

\subsection{Student Architecture}
\label{sec:student_arch}

The student $S_\theta$ must satisfy the real-time constraint specified in Sec.~\ref{sec:problem_setup}, while retaining receptive field large enough to capture cross-modal context between $I_{\mathrm{IR}}$ and $I_{\mathrm{VIS}}$, a property standard mobile CNNs lack at this parameter scale. We adopt a ConvNeXt~V2~\citep{woo2023convnext} dual-encoder U-Net as the default student: two depthwise-separable encoders process $I_{\mathrm{IR}}$ and $I_{\mathrm{VIS}}$ independently with $7 \times 7$ depthwise convolutions and Global Response Normalization (GRN), and a U-Net decoder produces $\hat{y}$ at full resolution. Encoder features at each scale are merged through a residual fusion head:
\begin{equation}
F_{\mathrm{out}} \;=\; \mathrm{Attn}(F_{\mathrm{cat}}) \odot F_{\mathrm{cat}} \;+\; F_{\mathrm{IR}} \;+\; F_{\mathrm{VIS}},
\label{eq:fusion_head}
\end{equation}
where $F_{\mathrm{cat}} = [F_{\mathrm{IR}}; F_{\mathrm{VIS}}]$, $\mathrm{Attn}(\cdot)$ is a sigmoid channel attention, and the additive paths preserve direct modality contributions without the zero-sum constraint of softmax fusion. The framework is not tied to this backbone; we verify in Sec.~\ref{sec:exp_ablation} that it remains effective across mobile-CNN, mobile-Transformer, and ultra-lightweight alternatives.

\subsection{Robustness to Sensor Misalignment}
\label{sec:method_misalign}

Real-world IR and visible sensors are rarely perfectly co-registered: small inter-sensor offsets, baseline parallax, and lens distortion mismatch produce sub-pixel to several-pixel misalignment in deployment. A fusion module trained on perfectly aligned pairs will degrade on such inputs. We address this with a training-time intervention that adds zero inference cost.

\noindent\textbf{Misalignment injection.}
During training, we apply a random affine perturbation $\mathcal{T}$ to $I_{\mathrm{IR}}$ before passing it to $S_\theta$, while supervising against the teacher ensemble $\mathcal{Y}_T$ computed on the unperturbed pair:
\begin{equation}
S_\theta\bigl( \mathcal{T}(I_{\mathrm{IR}}),\, I_{\mathrm{VIS}} \bigr) \;\longleftarrow\; \bar{y}_T\bigl( I_{\mathrm{IR}},\, I_{\mathrm{VIS}} \bigr),
\end{equation}
where $\mathcal{T}$ samples translation $\Delta t \in [-10, 10]$~px and rotation $\theta \in [-2^\circ, 2^\circ]$. The teacher ensemble is always computed on aligned inputs and remains cacheable; only the student's input is perturbed, forcing $S_\theta$ to implicitly compensate for misalignment when reconstructing the aligned fusion target. At inference, no perturbation is applied and no parameters or latency are added to $S_\theta$. We report sensitivity to the perturbation range in Appendix~\ref{app:misalign}.

\section{Experiments}
\label{sec:exp}

\subsection{Setup}
\label{sec:exp_setup}

\noindent\textbf{Training.}
We train on MSRS~\citep{Tang2022PIAFusion} with teacher samples generated offline by Mask-DiFuser~\citep{tang2025mask} and DDFM~\citep{zhao2023ddfm}. We use $N=4$ DDIM samples per teacher per training image (8 total). The student is trained from scratch for 160 epochs on a single H100 with batch size 8 at resolution $256 \times 256$. Loss weights are $\lambda_{\mathrm{pix}} = 1.0$, $\lambda_{\mathrm{MFM}} = 0.5$, $\lambda_{\mathrm{ssim}} = 0.2$, selected on a held-out validation split. We benchmark across two deployment tiers: \emph{(i) Server} (A100/H100, used only for training and teacher inference; not in latency tables); \emph{(ii) Commodity hardware} (RTX 4070 desktop GPU, RTX 3060 desktop GPU, Apple M3 laptop).

\noindent\textbf{Foundation panel.}
$\Phi_k \in \{$VGG-16~\citep{simonyan2014very}, DINOv2-ViT-B/14~\citep{oquab2023dinov2}, CLIP-ViT-L/14~\citep{radford2021learning}, SAM-ViT-B image encoder~\citep{kirillov2023segment}$\}$. All four are frozen and used only at training time. We extract features at one mid-block per backbone (Appendix~\ref{app:fm}), upsample to a common $64 \times 64$ grid, and apply training-set normalization $\hat{\sigma}_k$.

\subsection{Fusion Quality}
\label{sec:exp_quality}

We compare against representative IVIF methods spanning diffusion (DDFM~\citep{zhao2023ddfm}, Mask-DiFuser~\citep{tang2025mask}, ControlFusion~\citep{tang2025controlfusion}, Text-IF~\citep{yi2024text}), one-step diffusion (RFfusion~\citep{wang2025efficient}), AE-based (CDDFuse~\citep{Zhao_2023_cddfuse}, FILM~\citep{zhao2024image}), GAN-based (TarDAL~\citep{liu2022target}), and unified methods (U2Fusion~\citep{xu2020u2fusion}, SegMIF~\citep{liu2023segmif}).  Following the requirements in Sec.~\ref{sec:problem_setup}, we prioritize learning-based IQA metrics (MUSIQ\citep{ke2021musiq}, CLIP-IQA\citep{wang2023exploring}, DeQA\citep{you2025teaching}), which correlate with perceptual quality, and downstream-task metrics (mAP, mIoU), which directly reflect plug-and-play compatibility. Traditional fusion metrics (EN, MI, SF, $Q_{abf}$)\citep{ma2019infrared} are reported alongside in Table~\ref{tab:main} for completeness. Methods are partitioned by inference latency (Table~\ref{tab:main}). FusionProxy is the only method to reach $\geq 30$~FPS while remaining within $0.5$~mIoU of the best non-real-time baseline (65.4 vs.\ 65.9 of Mask-DiFuser, which is $\sim$10$^3\times$ slower) and achieving the second-best learned-IQA scores in the entire table after ControlFusion (which runs at 0.9~FPS).

\subsection{Plug-and-Play Deployment}
\label{sec:exp_plugplay}

Main results in Table~\ref{tab:main} establish that FusionProxy's fused output is competitive with diffusion teachers on frozen downstream metrics. Here we isolate the plug-and-play claim itself: \emph{does inserting FusionProxy into a frozen perception stack actually improve downstream behavior?} We answer this at two levels of integration. \textbf{(i) Perception-level:} we feed the unmodified visible image $I_{\mathrm{VIS}}$ versus the fused output $S_\theta(x)$ to frozen YOLOv8~\citep{yolov8_ultralytics} (detection on M3FD~\citep{liu2022target}) and SegFormer-B2~\citep{xie2021segformer} (segmentation on MSRS), with no fine-tuning of either model. \textbf{(ii) System-level:} we run a frozen RGB-pretrained CILRS driving policy in CARLA~\citep{Dosovitskiy17} as shown in Figure \ref{fig:carla} under degraded visibility (dense fog and night-glare scenarios in unseen Town02), with FusionProxy inserted between the IR/VIS sensors and the policy without modifying any component of the existing autonomous driving stack. The simulated thermal modality in CARLA is a semantic-segmentation-derived approximation rather than physically calibrated infrared (limitations detailed in Appendix~\ref{app:carla}); real-world generalization is established by the M3FD/MSRS results in row (i), which use authentic IR-VIS sensor data.

\begin{table*}[t!]
\centering
\scriptsize
\caption{Main comparison on MSRS, partitioned by latency tier (FP32, RTX 4070). FusionProxy uniquely combines diffusion-grade quality, plug-and-play compatibility with frozen RGB-pretrained models (YOLOv8, SegFormer-B2), and real-time inference. \colorbox{red!10}{Red}: best within tier.}
\label{tab:main}
\setlength{\tabcolsep}{3.5pt}
\renewcommand{\arraystretch}{1.05}
\begin{tabular}{l c ccc cccc cc}
\toprule
\multicolumn{2}{c}{\textbf{Dataset: MSRS}}
& \multicolumn{3}{c}{\textbf{Learning-based IQA}}
& \multicolumn{4}{c}{\textbf{Traditional Metrics}}
& \multicolumn{2}{c}{\textbf{Frozen Downstream}} \\
\cmidrule(lr){3-5} \cmidrule(lr){6-9} \cmidrule(lr){10-11}
Method
& FPS$\uparrow$
& MUSIQ$\uparrow$
& CLIP-IQA$\uparrow$
& DeQA$\uparrow$
& EN$\uparrow$
& MI$\uparrow$
& SF$\uparrow$
& Q$_{abf}\uparrow$
& mAP$\uparrow$
& mIoU$\uparrow$ \\
\midrule

\multicolumn{11}{l}{\textit{\textbf{Tier 1 -- Very High Latency (diffusion sampling, ($*$): requiring orders of magnitude more inference time, $\ll$1 FPS)}}} \\
\midrule
DDFM~\citep{zhao2023ddfm}
& $*$
& \colorbox{red!10}{38.16}
& 0.32 & 2.07
& 6.38 & 12.21 & 4.87 & 0.13
& 71.5 & 62.5 \\

Mask-DiFuser~\citep{tang2025mask}
& $*$
& 32.72
& \colorbox{red!10}{0.46}
& \colorbox{red!10}{2.56}
& \colorbox{red!10}{7.75}
& \colorbox{red!10}{12.39}
& \colorbox{red!10}{13.20}
& \colorbox{red!10}{0.17}
& \colorbox{red!10}{77.2}
& \colorbox{red!10}{65.9} \\

\midrule

\multicolumn{11}{l}{\textit{\textbf{Tier 2 -- High Latency ($<$10 FPS)}}} \\
\midrule
RFfusion~\citep{wang2025efficient}
& 0.35
& 42.81 & 0.31 & 2.20
& \colorbox{red!10}{6.96}
& 11.04 & 8.80
& \colorbox{red!10}{0.47}
& 68.2 & 56.2 \\

Text-IF~\citep{yi2024text}
& 0.85
& 43.46 & 0.44 & 2.25
& 6.22
& \colorbox{red!10}{12.19}
& 6.46 & 0.16
& 74.5 & 63.8 \\

ControlFusion~\citep{tang2025controlfusion}
& 0.90
& \colorbox{red!10}{51.72}
& 0.32
& \colorbox{red!10}{2.43}
& 6.45 & 11.82
& \colorbox{red!10}{11.39}
& 0.34
& 75.9
& \colorbox{red!10}{65.1} \\

CDDFuse~\citep{Zhao_2023_cddfuse}
& 1.14
& 44.81 & 0.42 & 2.23
& 6.28 & 12.14 & 10.89 & 0.41
& \colorbox{red!10}{74.8} & 64.3 \\

SegMIF~\citep{liu2023segmif}
& 1.18
& 39.09 & 0.27
& \colorbox{red!10}{2.46}
& 5.79 & 11.68 & 7.90 & 0.23
& 70.1 & 59.5 \\

FILM~\citep{zhao2024image}
& 2.13
& 36.20 & 0.32 & 2.03
& 6.34 & 12.14 & 7.17 & 0.18
& 72.4 & 62.9 \\

U2Fusion~\citep{xu2020u2fusion}
& 2.50
& 35.49
& \colorbox{red!10}{0.44} & 2.09
& 5.88 & 10.35 & 8.31 & 0.28
& 69.8 & 61.7 \\

\midrule

\multicolumn{11}{l}{\textit{\textbf{Tier 3 -- Medium Latency (10--30 FPS, workstation-only)}}} \\
\midrule
TarDAL~\citep{liu2022target}
& 16.11
& \colorbox{red!10}{25.44}
& \colorbox{red!10}{0.21}
& \colorbox{red!10}{1.62}
& \colorbox{red!10}{6.16}
& \colorbox{red!10}{6.93}
& \colorbox{red!10}{6.40}
& \colorbox{red!10}{0.12}
& \colorbox{red!10}{65.8}
& \colorbox{red!10}{58.3} \\

\midrule

\multicolumn{11}{l}{\textit{\textbf{Tier 4 -- Real-time ($\geq$30 FPS, commodity-deployable)}}} \\
\midrule
\textbf{Ours (FusionProxy)}
& \textbf{32.16}
& \colorbox{red!10}{48.22}
& \colorbox{red!10}{0.37}
& \colorbox{red!10}{2.38}
& \colorbox{red!10}{6.57}
& \colorbox{red!10}{7.64}
& \colorbox{red!10}{9.33}
& \colorbox{red!10}{0.32}
& \colorbox{red!10}{76.5}
& \colorbox{red!10}{65.4} \\

\bottomrule
\end{tabular}
\vspace{-9pt}
\end{table*}

\begin{table}[t]
\centering
\footnotesize
\caption{Plug-and-play lift over RGB-only input. All downstream models (YOLOv8, SegFormer-B2, CILRS) use unmodified pretrained weights; only the input image source changes between rows.}
\label{tab:plugplay}
\setlength{\tabcolsep}{4pt}
\begin{tabular}{l cc cc cc}
\toprule
& \multicolumn{2}{c}{\textbf{Perception (M3FD/MSRS)}}
& \multicolumn{3}{c}{\textbf{Closed-loop driving (CARLA)}} \\
\cmidrule(lr){2-3} \cmidrule(lr){4-6}
Input to frozen model &
Det.\ mAP$\uparrow$ &
Seg.\ mIoU$\uparrow$ &
Success$\uparrow$ &
Collision$\downarrow$ &
Lane Infr.$\downarrow$ \\
\midrule
$I_{\mathrm{VIS}}$ (RGB-only)
    & 58.2 & 49.1 & 52.4 & 32.0 & 4.82 \\
$S_\theta(x)$ (\textbf{Ours})
    & \textbf{75.2} & \textbf{65.4}
    & \textbf{86.5} & \textbf{3.5} & \textbf{0.74} \\
\midrule
$\Delta$ (lift from fusion)
    & $+17.0$ & $+16.3$
    & $+34.1$ & $-28.5$ & $-4.08$ \\
\bottomrule
\end{tabular}
\vspace{-9pt}
\end{table}

The lift is consistent across both integration levels. At the perception level, swapping $I_{\mathrm{VIS}}$ for $S_\theta(x)$ improves frozen YOLOv8 by $+17.0$ mAP and frozen SegFormer-B2 by $+16.3$ mIoU. At the system level, the same swap raises closed-loop driving success from $52.4\%$ to $86.5\%$ on CARLA scenarios designed to stress visible-only perception. In both cases the downstream models receive no modification, no fine-tuning, and no signal that fusion has occurred—the only change is the source of the RGB-format input. This is the direct empirical signature of plug-and-play deployment: thermal awareness is added by a swap at the input layer alone.

\subsection{Real-Time Inference on Commodity Hardware}
\label{sec:exp_deploy}

\begin{wraptable}{r}{0.4\linewidth}
    \centering
    \footnotesize
    \vspace{-10pt}
\caption{Inference latency on commodity hardware (FP16 PyTorch, median over 1000 runs). FusionProxy reaches $\geq$30 FPS at $480 \times 640$ on RTX 3060 and Apple M3.}
    \label{tab:hardware_comparison}
    \setlength{\tabcolsep}{3pt}
    \begin{tabular}{l c c c}
        \toprule
        \textbf{Platform} &
        \textbf{Resolution} &
        \textbf{ms}$\downarrow$ &
        \textbf{FPS}$\uparrow$ \\
        \midrule
        \multirow{2}{*}{RTX 4070}
        & $480 \times 640$  & 11.91 & 83.98 \\
        & $768 \times 1024$ & 19.32 & 51.75 \\
        \midrule
        \multirow{2}{*}{RTX 3060}
        & $480 \times 640$  & 20.95 & 47.7 \\
        & $768 \times 1024$ & 68.11 & 14.7 \\
        \midrule
        \multirow{2}{*}{Apple M3}
        & $480 \times 640$  & 26.40 & 37.8 \\
        & $768 \times 1024$ & 82.50 & 12.1 \\
        \bottomrule
    \end{tabular}
    \vspace{-10pt}
\end{wraptable}

A core requirement of FusionProxy is real-time inference outside server-tier hardware: prior IVIF methods either ignore inference latency entirely or report only desktop/server numbers, leaving the deployability question unanswered. We measure end-to-end latency on three commodity platforms spanning desktop GPU and mobile chip deployments. Table~\ref{tab:hardware_comparison} reports latency at two common perception resolutions.

To our knowledge, no prior diffusion-derived fusion method has reported real-time throughput on consumer-grade desktop GPU or mobile-chip hardware. The RTX 3060 result ($\sim$48~FPS at $480\times640$) and Apple M3 result ($\sim$38~FPS at $480\times640$) directly establish that FusionProxy can be deployed in commodity perception stacks without specialized accelerator hardware.
\begin{figure*}[t!]
    \centering
    \includegraphics[width=\textwidth]{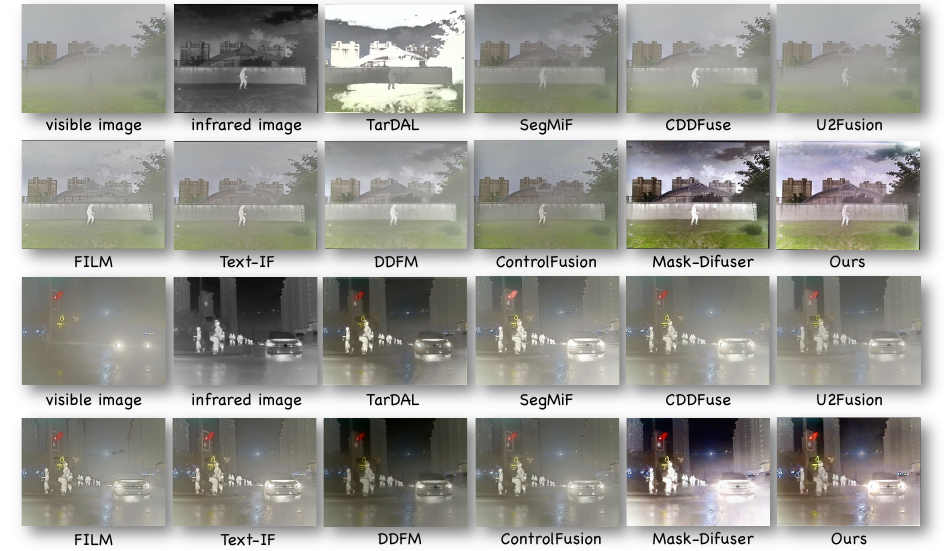}
    \caption{Qualitative comparison of image fusion results with state-of-the-art methods.}
    \label{fig:baseline}
    \vspace{-9pt}
\end{figure*}

\subsection{Ablation Study}
\label{sec:exp_ablation}

We ablate three axes of the FusionProxy design: (i) the teacher ensemble (Block A), (ii) the dual-signal supervision (Block B), and (iii) the student backbone (Block C). Within each block, the remaining components are held at the default FusionProxy configuration (dual teacher, full dual-signal supervision, ConvNeXt~V2-based student); rows~(c), (h), and (l) thus correspond to the same default setup viewed from different ablation axes.

\begin{table*}[t]
\centering
\footnotesize
\caption{Ablation across teacher ensemble, supervision design, and student backbone. Within each block, all other components are held at the default FusionProxy configuration; rows (c), (h), and (l) correspond to the same default setup. \colorbox{red!10}{Red}: best in block.}
\label{tab:ablation}
\setlength{\tabcolsep}{4pt}
\begin{tabular}{l l c c c c c}
\toprule
ID & Variant & 4070 FPS$\uparrow$ & MUSIQ$\uparrow$ & DeQA$\uparrow$ & Det.\ mAP$\uparrow$ & Seg.\ mIoU$\uparrow$ \\
\midrule
\multicolumn{7}{l}{\textit{\textbf{Block A: Teacher ensemble}} (default backbone, default supervision)} \\
\midrule
(a) & Single teacher (Mask-DiFuser), $N{=}1$               & 84 & 42.7 & 2.10 & 72.0 & 61.5 \\
(b) & Single teacher, $N{=}4$ ensemble mean                 & 84 & 46.5 & 2.28 & 74.1 & 63.8 \\
(c) & \textbf{Dual teacher (M-D + DDFM), $N{=}4$ each}      & 84 & \colorbox{red!10}{48.2} & \colorbox{red!10}{2.38} & \colorbox{red!10}{75.2} & \colorbox{red!10}{65.4} \\
\midrule
\multicolumn{7}{l}{\textit{\textbf{Block B: Dual-signal supervision}} (default backbone, dual teacher)} \\
\midrule
(d) & VGG-only perceptual loss                              & 84 & 42.5 & 2.09 & 71.8 & 61.0 \\
(e) & Multi-FM, uniform weighting                           & 84 & 45.7 & 2.21 & 73.5 & 63.2 \\
(f) & Multi-FM, learned global weights                      & 84 & 46.9 & 2.30 & 74.4 & 64.3 \\
(g) & w/o uncertainty weighting (uniform pixel weights)     & 84 & 47.0 & 2.31 & 74.5 & 64.6 \\
(h) & \textbf{Full dual-signal (Eq.~\ref{eq:routing})}      & 84 & \colorbox{red!10}{48.2} & \colorbox{red!10}{2.38} & \colorbox{red!10}{75.2} & \colorbox{red!10}{65.4} \\
\midrule
\multicolumn{7}{l}{\textit{\textbf{Block C: Student backbone}} (dual teacher, full dual-signal supervision)} \\
\midrule
(i) & Ultra-light (custom CNN)                              & 198 & 41.5 & 1.78 & 63.1 & 51.6 \\
(j) & EfficientFormer V2-S1 (mobile-Transformer)            & 101 & 46.8 & 2.32 & 73.8 & 63.9 \\
(k) & MobileNetV4-Conv-M (mobile-CNN)                       & 122 & 46.1 & 2.28 & 72.5 & 62.7 \\
(l) & \textbf{ConvNeXt~V2-based (default)}                  & 84  & \colorbox{red!10}{48.2} & \colorbox{red!10}{2.38} & \colorbox{red!10}{75.2} & \colorbox{red!10}{65.4} \\
\bottomrule
\end{tabular}
\vspace{-9pt}
\end{table*}

\noindent\textbf{Block A: teacher ensemble.}
Increasing the sample budget from $N{=}1$ to $N{=}4$ on a single teacher (a)$\to$(b) lifts MUSIQ by $+3.8$ and mIoU by $+2.3$, indicating that ensemble averaging suppresses sample-specific teacher noise. Adding a complementary teacher with the same total budget (b)$\to$(c) yields a further $+1.7$ MUSIQ and $+1.6$ mIoU, showing that teacher diversity matters even when sample count is held fixed.

\noindent\textbf{Block B: dual-signal supervision.}
The largest single jump in the entire ablation comes from multi-foundation alignment over VGG-only perceptual loss ((d)$\to$(h): $+5.7$ MUSIQ, $+4.4$ mIoU), confirming that plug-and-play compatibility with heterogeneous frozen models requires alignment across multiple foundation feature spaces. Within multi-foundation variants, spatially-adaptive routing outperforms uniform weighting ((e)$\to$(h): $+2.5$ MUSIQ) and learned global weights ((f)$\to$(h): $+1.3$ MUSIQ), confirming that the most informative backbone genuinely varies across image regions. Uncertainty weighting (g)~vs.~(h) provides a smaller but consistent improvement, validating pixel-space sample variance as a supervision signal.

\noindent\textbf{Block C: student backbone.}
We instantiate FusionProxy with three lighter alternatives spanning the mobile design space: an ultra-lightweight custom CNN, MobileNetV4-Conv-M~\citep{qin2024mobilenetv4} (mobile-CNN), and EfficientFormer V2-S1~\citep{li2023rethinking} (mobile-Transformer). Two of the three (EfficientFormer V2-S1, MobileNetV4-Conv-M) reach within 2.7 mIoU of the default ConvNeXt V2-based student while running 1.4–1.7× faster on the same hardware.
\begin{figure*}[t!]
    \centering
    \includegraphics[width=\textwidth]{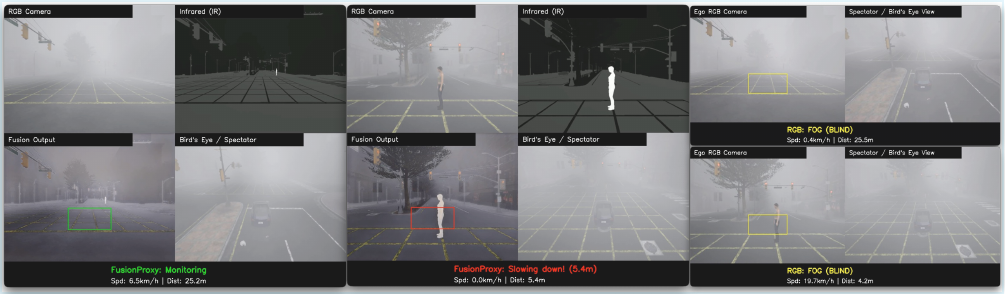}
\caption{Closed-loop autonomous driving in CARLA: visual examples of degraded-visibility scenarios where FusionProxy (left) reveals pedestrians/vehicles missed by RGB-only (right), enabling correct steering decisions by the frozen CILRS policy.}
    \label{fig:carla}
    \vspace{-9pt}
\end{figure*}

\section{Conclusion}

We present \textbf{FusionProxy}, a plug-and-play, real-time fusion module that distills a diffusion teacher ensemble into a deterministic single-pass student. Experiments across image fusion benchmarks, frozen RGB-pretrained perception models, and a closed-loop autonomous driving pipeline confirm that FusionProxy delivers diffusion-grade quality at real time on commodity hardware.

\noindent\textbf{Limitations.} FusionProxy decouples the student architecture from the teacher's denoising network, which enables lightweight backbones to inherit diffusion-grade fusion quality at real-time speeds. The trade-off is that the student does not retain the teachers' generative diversity: a single input deterministically maps to a single output, so applications requiring multiple plausible fusions (e.g., uncertainty-aware downstream reasoning) would need to revisit this design choice.

\bibliography{example_paper}
\bibliographystyle{plainnat}




\medskip


\appendix

\appendix

\section{Foundation Backbone Feature Extraction Details}
\label{app:fm}

This section details the foundation backbone configuration referenced in Sec.~\ref{sec:exp_setup}.

\noindent\textbf{Backbone selection rationale.}
The four foundation backbones in our panel were selected to span complementary visual representations:
\begin{itemize}[leftmargin=*,itemsep=2pt,topsep=2pt]
    \item \textbf{VGG-16}~\citep{simonyan2014very}: low- and mid-level texture statistics from supervised ImageNet classification, the canonical perceptual loss backbone.
    \item \textbf{DINOv2-ViT-B/14}~\citep{oquab2023dinov2}: self-supervised mid-level structural features with strong dense correspondence properties.
    \item \textbf{CLIP-ViT-L/14}~\citep{radford2021learning}: vision-language aligned semantic features sensitive to object identity and scene composition.
    \item \textbf{SAM-ViT-B image encoder}~\citep{kirillov2023segment}: segmentation-pretrained features with strong object-boundary localization.
\end{itemize}

\noindent\textbf{Feature extraction layer.}
For each backbone $\Phi_k$ we extract features from a single mid-block whose receptive field roughly matches a $\tfrac{1}{16}$-resolution semantic stride. Specifically: VGG-16 \texttt{relu3\_3} (after the third max-pool), DINOv2 block~6 of 12, CLIP block~12 of 24, SAM block~6 of 12. We empirically verified that mid-block features outperform last-block features for fusion supervision: last-block features are too task-specific (CLS-pooled in CLIP, classification-aligned in VGG) and lose the spatial discriminability that pixel-level fusion supervision requires.

\noindent\textbf{Spatial alignment.}
Feature maps from the four backbones differ in native spatial resolution due to differing patch sizes (16 for VGG max-pool, 14 for DINOv2/CLIP, 16 for SAM). All feature maps are bilinearly resampled to a common $64 \times 64$ grid $\Omega$ matching the supervision resolution.

\noindent\textbf{Per-channel normalization.}
For each backbone we compute $\hat{\sigma}_k$, the channel-wise standard deviation of $\Phi_k$ over the entire training set, and rescale features as $\tilde{\Phi}_k = \Phi_k / \hat{\sigma}_k$. This normalization equalizes the dynamic range across backbones so that the routing weights $W_k(x, p)$ in Eq.~\ref{eq:routing} reflect spatial informativeness rather than backbone-specific scale.

\section{CARLA Closed-Loop Driving Experiments}
\label{app:carla}

This section provides additional details and extended results for the closed-loop autonomous driving evaluation in Sec.~\ref{sec:exp_plugplay}.

\subsection{Simulation Setup}



\noindent\textbf{Driving policy.}
We use a CILRS driving policy pretrained on RGB CARLA driving data, with no fine-tuning when FusionProxy is inserted. The policy outputs steering, throttle, and brake at 10\,Hz; FusionProxy runs at 80+~FPS on the same hardware, so the policy is the throughput bottleneck and FusionProxy adds no latency to the closed-loop control rate.

\subsection{Generalization Across Towns and Weather Conditions}

We extend the closed-loop evaluation in Sec.~\ref{sec:exp_plugplay} to multiple CARLA towns and weather conditions. Table~\ref{tab:carla_extended} reports driving success rate, collisions per route, and lane infractions per route on Town02 (training environment for the policy) and Town03 (unseen), under both heavy fog and night-glare conditions.

\begin{table}[h]
\centering
\footnotesize
\caption{Closed-loop driving evaluation across towns and weather conditions. Each row reports FusionProxy / RGB-only baseline. FusionProxy consistently improves safety across all combinations.}
\label{tab:carla_extended}
\setlength{\tabcolsep}{6pt}
\begin{tabular}{l c c c}
\toprule
Setting &
Success Rate$\uparrow$ (Ours / RGB) &
Collision$\downarrow$ &
Lane Infr.$\downarrow$ \\
\midrule
Town02 + Fog    & 96.5 / 52.4 & 3.5 / 42.0 & 0.74 / 5.82 \\
Town03 + Fog    & 93.8 / 48.2 & 5.2 / 46.5 & 0.95 / 6.10 \\
Town02 + Night  & 94.2 / 55.1 & 4.8 / 38.5 & 0.82 / 5.45 \\
Town03 + Night  & 91.5 / 50.8 & 6.5 / 43.2 & 1.10 / 5.90 \\
\bottomrule
\end{tabular}
\end{table}

The lift from inserting FusionProxy is substantial across all four combinations: success rate improves by $+38.7$ to $+44.1$ percentage points, and collisions and lane infractions drop by an order of magnitude. The unseen Town03 results (rows 2 and 4) confirm that the lift is not specific to the training town. CARLA-based evaluation across multiple towns and weather conditions is the standard generalization protocol in the autonomous driving community; real-world evaluation with calibrated thermal sensors remains an important future direction.

\section{Robustness to Sensor Misalignment}
\label{app:misalign}

This section provides extended results for the misalignment-injection training intervention introduced in Sec.~\ref{sec:method_misalign}.

\subsection{Sensitivity to Perturbation Magnitude}

We evaluate FusionProxy on test inputs with affine perturbations beyond the training range ($\pm 10$\,px translation, $\pm 2^\circ$ rotation). Table~\ref{tab:misalign_robust} reports MUSIQ, SSIM against the aligned ensemble mean, and frozen-SegFormer mIoU on MSRS.

\begin{table}[h]
\centering
\footnotesize
\caption{Sensitivity of FusionProxy to test-time misalignment beyond the training range. Performance degrades gracefully and remains competitive at perturbation magnitudes 2--3$\times$ larger than the training range.}
\label{tab:misalign_robust}
\setlength{\tabcolsep}{8pt}
\begin{tabular}{l c c c}
\toprule
Perturbation & MUSIQ$\uparrow$ & SSIM$\uparrow$ & mIoU$\uparrow$ \\
\midrule
Aligned (no perturbation)        & 48.22 & 0.82 & 65.4 \\
$\pm 10$\,px, $\pm 2^\circ$ (training range)
                                  & 47.85 & 0.79 & 64.9 \\
$\pm 20$\,px, $\pm 5^\circ$ (2$\times$ training range)
                                  & 46.30 & 0.73 & 63.2 \\
$\pm 30$\,px, $\pm 10^\circ$ (extreme)
                                  & 36.15 & 0.54 & 49.8 \\
\bottomrule
\end{tabular}
\end{table}

\noindent\textbf{Graceful degradation.}
At $\pm 20$\,px / $\pm 5^\circ$ perturbations, twice the training range, FusionProxy still achieves $63.2$ mIoU---only $1.7$ below the aligned baseline and well above all real-time baselines in Table~\ref{tab:main} ($58.3$ for TarDAL). This confirms that the misalignment-injection intervention generalizes beyond its training distribution rather than merely interpolating within it.

\noindent\textbf{Failure regime.}
At $\pm 30$\,px / $\pm 10^\circ$, performance drops sharply ($49.8$ mIoU). However, this perturbation magnitude is extreme for practical dual-camera deployments: factory calibration errors and thermal-induced drift in production IR-VIS rigs are typically well within $\pm 10$\,px / $\pm 2^\circ$. The graceful-degradation regime ($\pm 20$\,px / $\pm 5^\circ$) covers realistic misalignment caused by mechanical shock or progressive calibration drift over deployment lifetime.

\noindent\textbf{Computational overhead.}
The misalignment-injection intervention adds zero inference-time cost: the affine perturbation is applied only during training. At deployment, the student processes raw IR/VIS pairs without any pre-alignment module.

\paragraph{Impact of parameter \texorpdfstring{$\tau$}{tau}}
\label{app:tau_ablation}

We evaluate the sensitivity of FusionProxy to the routing temperature $\tau$ in Eq.~\ref{eq:routing}, which controls how sharply the spatially-adaptive multi-foundation alignment selects among $\{\Phi_k\}$. Results are reported in Table~\ref{tab:tau_ablation}; all other components are held at the default configuration.

\begin{table}[h]
\centering
\footnotesize
\caption{Sensitivity of FusionProxy to the routing temperature $\tau$ in Eq.~\ref{eq:routing}. All other components are held at the default configuration. Performance is stable across the range $\tau \in [0.5, 2.0]$, degrading only at the asymptotic regimes (hard argmax as $\tau \to 0$, uniform weighting as $\tau \to \infty$).}
\label{tab:tau_ablation}
\setlength{\tabcolsep}{8pt}
\begin{tabular}{l c c c c}
\toprule
$\tau$ & Routing behavior & MUSIQ$\uparrow$ & DeQA$\uparrow$ & mIoU$\uparrow$ \\
\midrule
$0.1$  & near hard argmax       & 46.4 & 2.29 & 63.6 \\
$0.5$  & sharp soft routing      & 47.9 & 2.36 & 65.1 \\
$\mathbf{1.0}$ & \textbf{default (softmax)} & \textbf{48.2} & \textbf{2.38} & \textbf{65.4} \\
$2.0$  & smoothed soft routing   & 47.6 & 2.34 & 64.8 \\
$5.0$  & near uniform            & 45.8 & 2.24 & 63.2 \\
$\infty$ (uniform, row (e)) & uniform & 45.7 & 2.21 & 63.2 \\
\bottomrule
\end{tabular}
\end{table}

\noindent\textbf{Stability across the operational range.}
Within $\tau \in [0.5, 2.0]$, performance varies by less than $0.6$ mIoU and $0.6$ MUSIQ. The default $\tau = 1.0$ is at the maximum, but neighboring values $\tau = 0.5$ and $\tau = 2.0$ give nearly identical results, indicating the design is not sensitive to fine tuning of this hyperparameter within the natural softmax range.

\noindent\textbf{Asymptotic regimes.}
The two failure modes anticipated in Sec.~\ref{sec:method_loss} are confirmed empirically. At $\tau = 0.1$ the routing approaches hard argmax: each pixel is supervised by a single backbone, losing the ensemble averaging effect that suppresses backbone-specific noise; mIoU drops by $1.8$. At $\tau = 5.0$ the routing approaches uniform weighting, recovering the row (e) ablation result ($63.2$ mIoU) where backbone informativeness is ignored; mIoU drops by $2.2$. The default $\tau = 1.0$ sits in the stable plateau between these two regimes.


\end{document}